\pdfoutput=1
\documentclass[10pt,logo,copyright]{nvidiatechreport}
\usepackage[square,numbers]{natbib}

\usepackage[utf8]{inputenc} %
\usepackage[T1]{fontenc}    %

\usepackage{parskip}        %
\usepackage{url}            %
\usepackage{hyperref}       %
\usepackage{booktabs}       %
\usepackage{amsfonts}       %
\usepackage{nicefrac}       %
\usepackage{microtype}      %
\usepackage{xcolor}         %
\usepackage[dvipsnames]{xcolor} %
\usepackage{graphicx}
\usepackage{animate}        %
\usepackage{subcaption}
\usepackage{tabularx}
\usepackage{makecell}
\usepackage{adjustbox}
\usepackage{setspace}
\newcolumntype{M}[1]{>{\centering\arraybackslash}m{#1}}
\usepackage{float}
\usepackage{tikz}
\usetikzlibrary{positioning,shapes,arrows}
\usepackage{amsmath,amsfonts,bm, bbm,leftindex}
\usepackage{multirow}
\usepackage{comment}
\usepackage{gensymb}
\usepackage{lipsum}
\usetikzlibrary{arrows.meta, positioning, fit}
\usepackage[para]{threeparttable}
\usepackage{tikz}
\usetikzlibrary{tikzmark}

\definecolor{darkred}{rgb}{0.7, 0.0, 0.0}

\usepackage{pifont}
\usepackage{wrapfig}

\usepackage[nameinlink]{cleveref}
\crefname{equation}{Eq.}{Eqs.}
\crefname{figure}{Fig.}{Figs.}
\crefname{section}{Sec.}{Sec.}
\crefname{appendix}{App.}{App.}
\crefname{table}{Tab.}{Tabs.}
\crefname{algorithm}{Algo}{Algo}
\crefname{thm}{Thm}{Thm}
\Crefname{thm}{Thm}{Thm}
\crefname{prop}{Prop}{Prop}
\usepackage{longtable}

\usepackage{amsfonts}
\usepackage{bm}
\usepackage{amsmath}
\usepackage{mathtools}
\usepackage{multirow}
\usepackage{booktabs}
\usepackage{caption}
\usepackage{enumitem}
\usepackage{wrapfig,lipsum,booktabs}
\usepackage{caption}
\usepackage{bbm}
\usepackage{multirow}
\usepackage{pifont}
\usepackage[linesnumbered,ruled,vlined]{algorithm2e}

\SetCommentSty{mycommfont}
\SetAlCapFnt{\small} 
\usepackage{etoc}
\SetAlgorithmName{Algo}{Algo}{}

\renewcommand{\paragraph}[1]{{\vspace{1mm}\noindent \bf #1}.}

\newcommand{\bx}{\mathbf{x}}
\newcommand{\bxhat}{\mathbf{\hat{x}}}
\newcommand{\bc}{\mathbf{c}}

\newcommand{\br}{\mathbf{r}}
\newcommand{\bj}{\mathbf{j}}
\newcommand{\beff}{\mathbf{f}}
\newcommand{\bmask}{\mathbf{m}}
\newcommand{\bb}{\mathbf{b}}

\newcommand{\reals}{\mathbb{R}}

\newcommand{\link}{https://research.nvidia.com/labs/sil/projects/kimodo}

\newcommand{\model}{{{Kimodo}}\xspace}
\newcommand{\bodymodel}{{{SOMA}}\xspace}

\newcommand{\crefnames}[3]{%
  \@for\next:=#1\do{%
    \expandafter\crefname\expandafter{\next}{#2}{#3}%
  }%
}

\title{Kimodo: Scaling Controllable Human Motion Generation}

\begin{document}

\author{
    Davis Rempe$^\ast$,
    Mathis Petrovich$^\ast$,
    Ye Yuan,
    Haotian Zhang,
    Xue Bin Peng,
    Yifeng Jiang,
    Tingwu Wang,
    Umar Iqbal, 
    David Minor,
    Michael de Ruyter,
    Jiefeng Li,
    Chen Tessler,
    Edy Lim, 
    Eugene Jeong, 
    Sam Wu, 
    Ehsan Hassani,
    Michael Huang,
    Jin-Bey Yu,
    Chaeyeon Chung,
    Lina Song, 
    Olivier Dionne,
    Jan Kautz,
    Simon Yuen,
    Sanja Fidler
    \\
	\small NVIDIA \\
    \small$^\ast$ Co-First Authors \\
    \small \tt{\href{\link}{\link}}
}

\maketitle
\newcommand{\todo}[1]{\textcolor{red}{ [TODO: \textbf{#1}]}}
\newcommand{\davis}[1]{\textcolor{green}{ [Davis: \textbf{#1}]}}
\newcommand{\mathis}[1]{\textcolor{teal}{ [Mathis: \textbf{#1}]}}
\newcommand{\jason}[1]{\textcolor{blue}{ [Jason: \textbf{#1}]}}

\begin{abstract} 
High-quality human motion data is becoming increasingly important for applications in robotics, simulation, and entertainment. Recent generative models offer a potential data source, enabling human motion synthesis through intuitive inputs like text prompts or kinematic constraints on poses. However, the small scale of public mocap datasets has limited the motion quality, control accuracy, and generalization of these models. In this work, we introduce Kimodo, an expressive and controllable kinematic motion diffusion model trained on 700 hours of optical motion capture data. Our model generates high-quality motions while being easily controlled through text and a comprehensive suite of kinematic constraints including full-body keyframes, sparse joint positions/rotations, 2D waypoints, and dense 2D paths. This is enabled through a carefully designed motion representation and two-stage denoiser architecture that decomposes root and body prediction to minimize motion artifacts while allowing for flexible constraint conditioning. Experiments on the large-scale mocap dataset justify key design decisions and analyze how the scaling of dataset size and model size affect performance.
\end{abstract}

\abscontent

\section{Introduction}
While human motion data has always been central to games and other media, recent advances in robotics and physical AI have increased the demand for such data.
In robotics, human demonstrations allow humanoids to move realistically and complete complex tasks~\cite{tessler2024maskedmimic,tessler2025maskedmanipulator,luo2025sonic}. 
Digital twins and industrial simulations need dynamic humans to populate and interact with environments. 
And in established domains such as game development, the growing scope of interactive experiences necessitates plausible digital humans at scale.

Obtaining high-quality 3D human motion data is challenging, though. 
Traditional hand animation is often tedious and requires substantial domain expertise, while studio motion capture (mocap) is expensive and requires heavy instrumentation of both the actors and environments. 
Teleoperation has become a popular method to collect demonstrations directly with robots~\cite{ze2025twist}, but this process is slow and results in awkward, unnatural behaviors. 
While videos offer a rich source of human motions~\cite{li2025genmo}, recovering high-quality 3D motions from monocular videos remains a challenging research problem.

In this work, we contend that generative models offer an alternative motion data acquisition paradigm, which can easily synthesize high-quality human motion data with precise control.
For such a model to be useful in practice, it should maintain the advantages of traditional motion acquisition approaches while increasing accessibility for novice animators (\eg, roboticists). In particular, an ideal model should: (1) produce high-quality motions on par with optical mocap, (2) provide a versatile and directable interface akin to hand animation, but with more intuitive controls, and (3) be able to generate a diverse corpus of motions to support a large variety of applications.

\begin{figure}[t]
    \centering
    \includegraphics[width=1.0\linewidth]{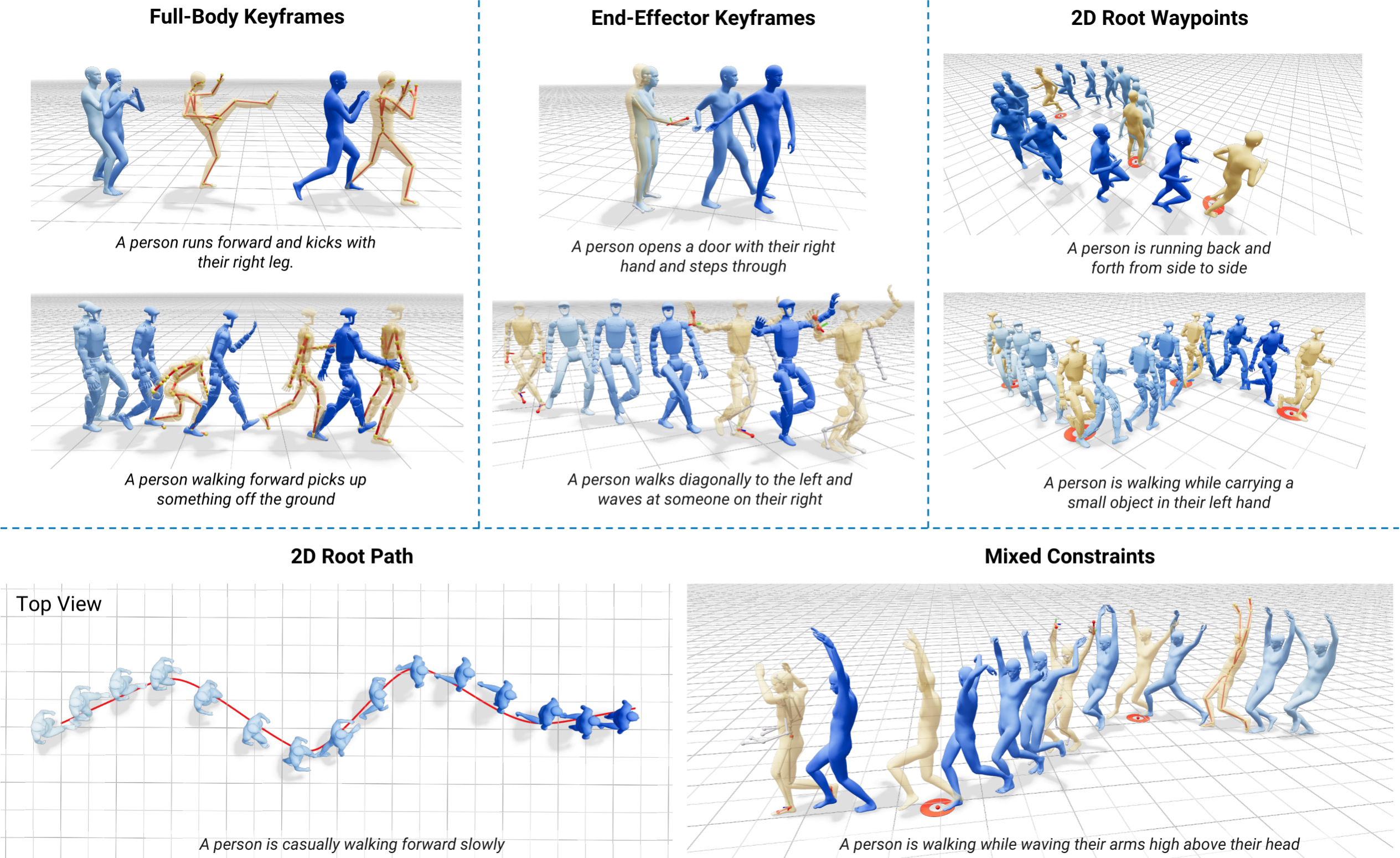}
    \caption{\textbf{Controllable Motion Generation.} Kimodo supports flexible and intuitive control for motion generation through text prompting combined with an extensive suite of kinematic constraints. By training on 700 hours of optical mocap data, the model achieves precise control accuracy for a large variety of behaviors. In each example, constrained joints are indicated with a red color, and generated poses at constrained frames are highlighted in yellow. Time progression is indicated by lighter to darker blue coloring.}
    \label{fig:constraint_results}
\end{figure}

Recently, rapid progress has been made in human motion generation. 
Modern generative models such as diffusion~\cite{tevet2023mdm,zhang2024motiondiffuse}, masked models~\cite{guo2024momask}, and tokenized transformers~\cite{zhang2023generating} have enabled intuitive motion generation by conditioning on text prompts.
Some approaches also support \emph{control} over generation through various kinematic constraints such as keyframe poses, 2D waypoints, and joint trajectories~\cite{guo2024momask, karunratanakul2023gmd, xie2024omnicontrol, cohan2024condmdi}.
These works show promising results and have inspired a wave of new research, but their motion quality, control accuracy, and generalization ability are limited by relatively small publicly available datasets~\cite{guo2022humanml3d,Plappert2016}. Furthermore, saturated benchmarks on these datasets make it difficult to differentiate which design decisions are critical to achieve effective modeling.
Several works attempt to scale up human motion generation models and improve generalization by training on large volumes of motion data recovered from videos~\cite{hymotion2025,wang2025scaling,lu2025scamo,lin2023motion}, but these approaches tend to compromise motion quality due to the inherent inaccuracies in video reconstruction.

In this report, we introduce \textbf{\model}, a \textbf{ki}nematic \textbf{mo}tion \textbf{d}iffusi\textbf{o}n model that 
trains at scale to enable intuitive authoring of high-quality motions through text prompting and an extensive set of kinematic constraints. 
Our model uses a carefully designed motion representation and two-stage diffusion architecture that decomposes root and body motion to accurately follow user specifications while minimizing common artifacts, such as floating and foot skating.
In addition to text inputs, the model natively supports a suite of constraints on generated poses, including full-body keyframes, sparse joint positions/rotations, 2D waypoints, 2D path following, and foot contact patterns.  
A key component to effectively train \model is the Bones Rigplay~\cite{bones_ai_datasets} dataset, a large studio mocap dataset containing 700 hours of production-quality human motion with corresponding text descriptions. This large dataset also enables a comprehensive evaluation of various design decisions on a wide range of behaviors and scenarios.

To facilitate motion generation for robotics, simulation, games, and other applications, we have released model checkpoints for \model trained on the SOMA body model~\cite{saito2026soma} and the Unitree G1 robot~\cite{unitree2024g1}. Along with these models, we have included generation code and an interactive interface for motion authoring (see \cref{fig:authoring_demo}) to demonstrate the model's full capabilities. 
In the remainder of this report, \cref{sec:key_results} shows the key capabilities of our model, which enable practical authoring of human motions. 
Next, \cref{sec:dataset} describes the Bones Rigplay dataset and \cref{sec:method} details the core design of our two-stage transformer denoiser along with the training recipe. 
\cref{sec:relwork} positions our work in the context of prior systems, and lastly, \cref{sec:experiments} provides a comprehensive evaluation of key design decisions in our model and explores how scaling dataset size, model size, and batch size (GPUs) impacts model performance.
\section{Key Results}
\label{sec:key_results}

\begin{figure}[t]
    \centering
    \includegraphics[width=1.0\linewidth]{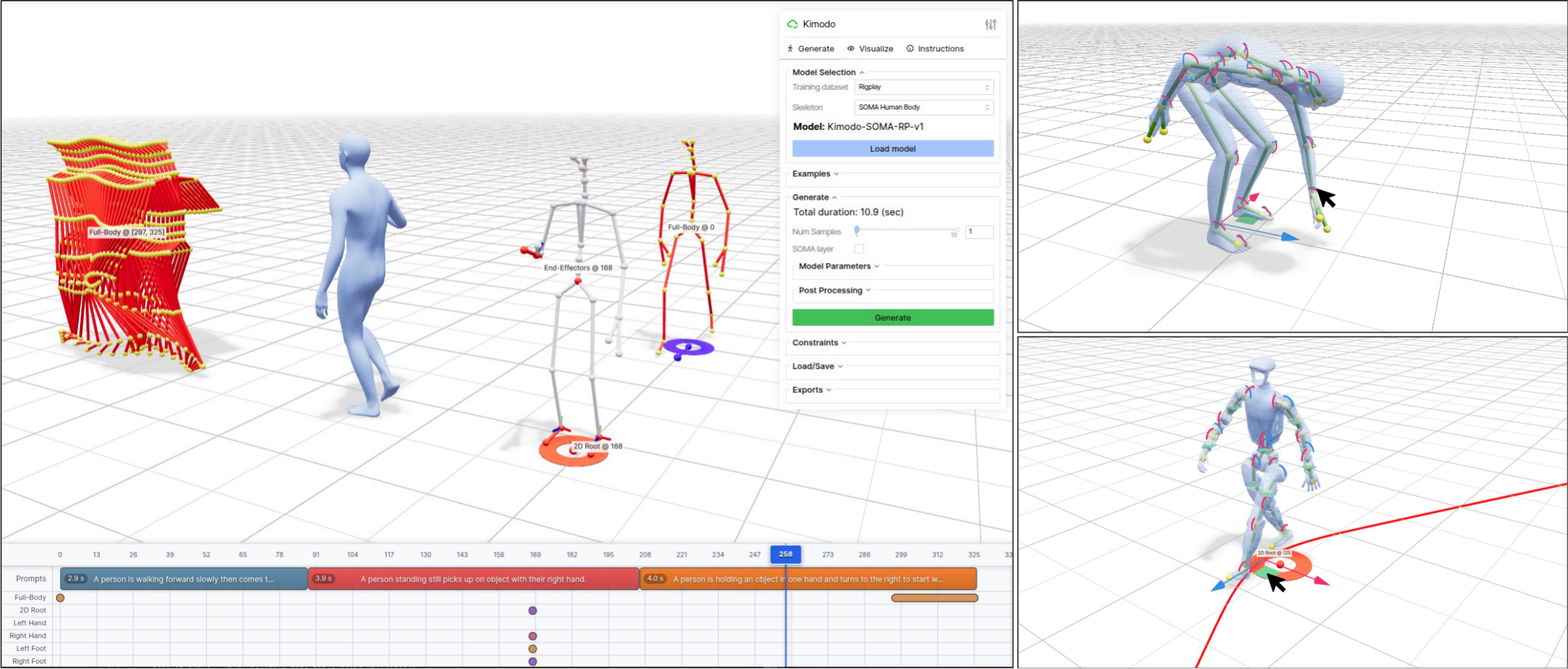}
    \caption{\textbf{Motion Authoring Demo.} (Left) Our authoring interface built with Viser~\cite{yi2025viser} allows intuitive control over \model for motion generation. The timeline panel allows users to specify text prompts and constraints at specific frames or intervals, which are displayed in the 3D viewer. The options panel on the right side of the interface controls various generation parameters. (Right) In editing mode, users have fine-grained control to pose and translate the character at constrained frames. Editing and generation can be done on either the SOMA body skeleton~\cite{saito2026soma} or Unitree G1 robot.}
    \label{fig:authoring_demo}
\end{figure}

\model is designed for intuitive authoring of high-quality human motions. 
To showcase its capabilities, we developed an interactive demo with Viser~\cite{yi2025viser}, which allows a user to generate motions from a combination of text prompts and kinematic constraints (see \cref{fig:authoring_demo}). 
In this section, we detail the key features of the model within this demo interface, explore how scaling affects the model performance, and demonstrate downstream use of the model for humanoid robotics. 
The motions produced by our model are best viewed in the supplementary videos on the project webpage (\hyperlink{https://research.nvidia.com/labs/sil/projects/kimodo/}{https://research.nvidia.com/labs/sil/projects/kimodo/}) or by running the demo in the released codebase (\hyperlink{https://github.com/nv-tlabs/kimodo}{https://github.com/nv-tlabs/kimodo}).

\subsection{Motion Generation with Text Control}
\label{sec:key_results:text}

\begin{figure}[t!]
    \centering
    \includegraphics[width=1.0\linewidth]{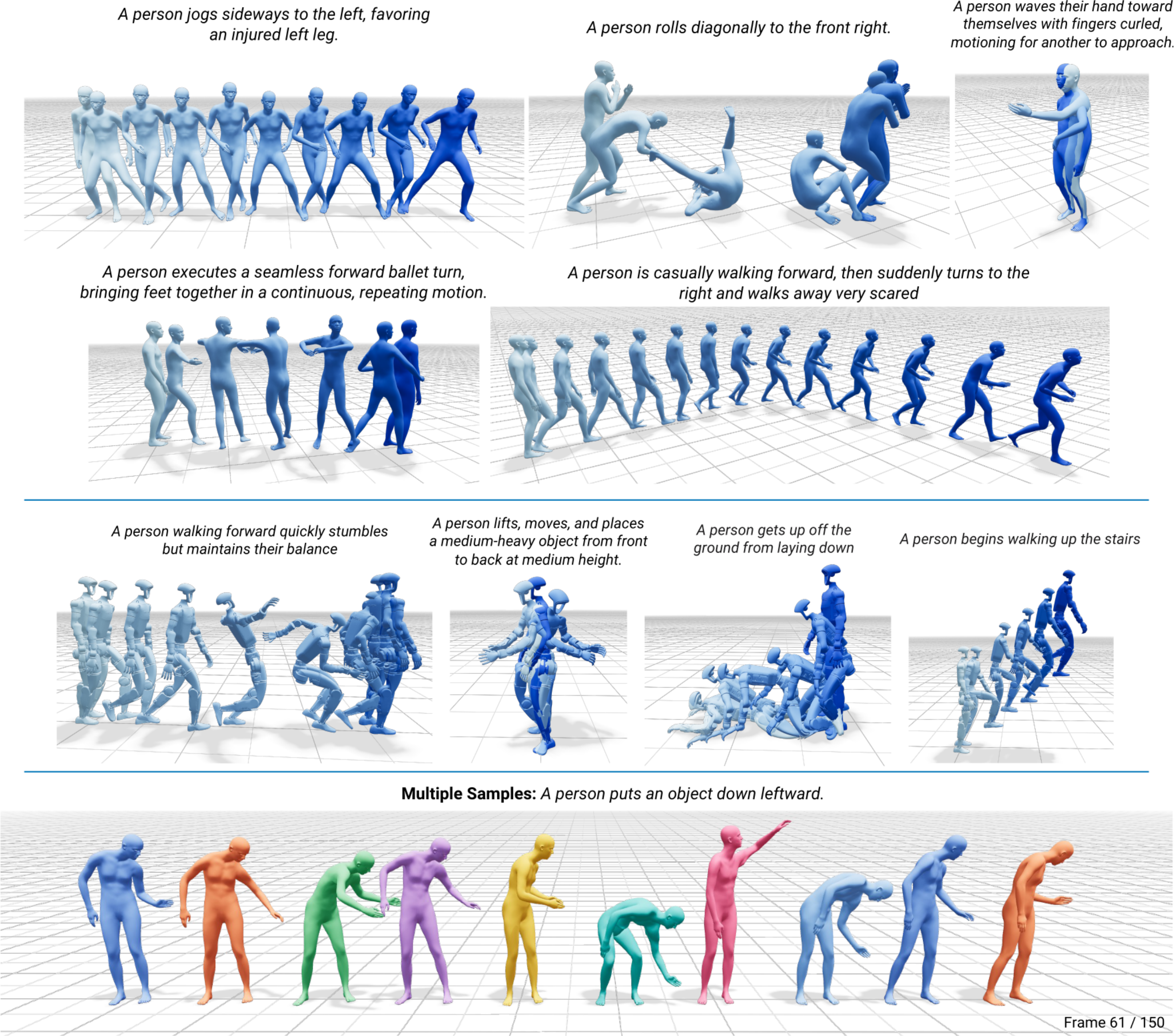}
    \caption{\textbf{Text-to-Motion Results.} (Top) \model enables generating high-quality human motions for a variety of behaviors on the SOMA body skeleton. Time progression is indicated by lighter to darker blue coloring. (Middle) Motions can also be generated directly on the G1 robot to easily collect plausible demonstrations. (Bottom) The same frame is visualized from ten different generated motion samples for the same prompt, demonstrating the diversity of \model outputs.}
    \label{fig:text2motion_results}
\end{figure}

As shown in \cref{fig:text2motion_results}, motion generation with \model can be easily controlled through intuitive text prompting. 
Our model, trained on the SOMA body skeleton~\cite{saito2026soma} at 30 fps, enables generating realistic human motions for a variety of behaviors contained in the training data, including locomotion, everyday activities, dancing, actions and combat, and different motion styles. The model can handle potentially complex text descriptions that contain multiple actions performed in sequence (``A person walks forward \textit{then} waves their arms.") or simultaneously (``A person walks forward \textit{while} waving their arms.").
Additionally, we retargeted the training dataset to the Unitree G1 robot using the SOMA retargeter~\cite{soma_retarget}, allowing direct generation of kinematic robot demonstrations for important skills like recovering from stumbles and falls, and object interactions. 
The bottom row of \cref{fig:text2motion_results} demonstrates the diversity of generated motions. The same frame from ten samples from the model is visualized, each with a different color. 
Though all samples use the same text prompt describing a put-down motion, the model generates plausible variations including one and two-handed, high and low, and variable timing. 

\begin{wrapfigure}{r}{0.45\textwidth} %
\vspace{-2mm}
    \centering
    \includegraphics[width=\linewidth]{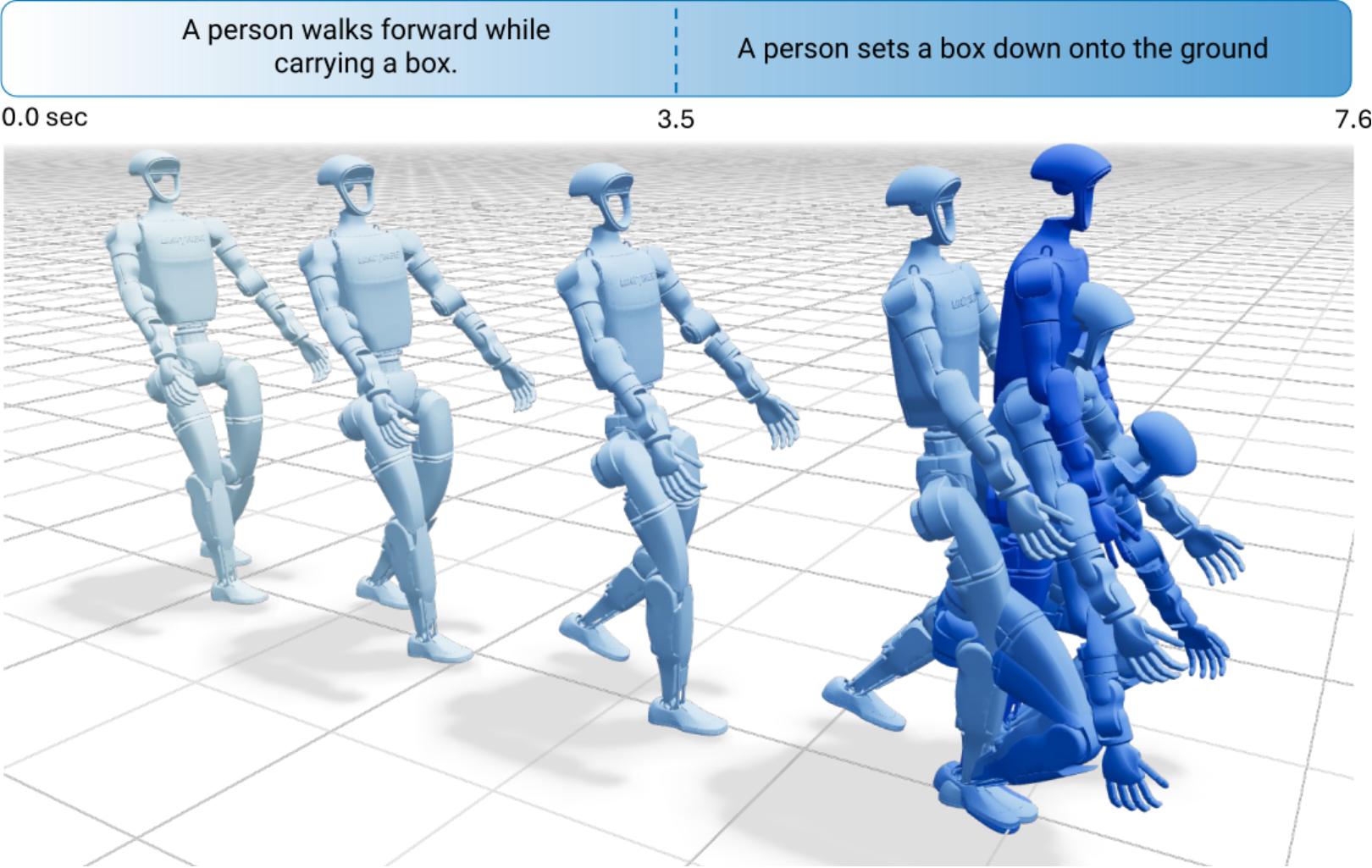}
    \caption{\textbf{Multi-Prompt Generation.} Longer motion sequences can be generated from multiple prompts with the demo's timeline interface. Motions are generated sequentially with constraints between them for continuity.}
    \vspace{-9mm}
    \label{fig:multiprompt_results}
\end{wrapfigure}

\model is trained to take a single prompt as input and expects the prompt to start with the subject, such as ''A person...", ``An old person...", or ``A zombie...". However, our interactive authoring demo enables chaining together multiple prompts in sequence with plausible transitions. 
As shown in \cref{fig:multiprompt_results}, using multiple prompts can be effective for performing a sequence of actions that the model may struggle with when specified as a single prompt. Please see \cref{sec:method:inference} for technical details of multi-prompt generation.

\model is designed as an offline motion authoring model, and is best suited for generating one or more motions in a batched fashion. On an NVIDIA RTX 3090, generating a motion from a single prompt takes anywhere from 2 to 5 sec, depending on the specified duration. The model is trained on a maximum of 10 sec sequences.

\subsection{Kinematic Control with Pose Constraints}
\label{sec:key_results:constraints}

Precise control can be achieved through kinematic constraints on generated poses.
As shown in \cref{fig:constraint_results}, \model supports a wide range of constraint types. 
\textit{Full-body keyframes} constrain all the joint positions of the character at specific frames. 
\textit{End-effector keyframes} specify the position and rotation of one or more hand or foot joints. 
For 2D root constraints, \textit{waypoints} specify sparse targets on the ground for the character to hit, or a full dense \textit{path} can be used to constrain an interval of motion. 
Constraints can be mixed arbitrarily to achieve desired motions.

Kinematic control enables several useful motion authoring applications. Animators can use \model to generate plausible transitions between existing mocap clips through in-betweening, with full-body constraints placed at the start and end of the generated sequence. For characters to realistically move about an environment, 2D paths or waypoints can be specified automatically using traditional animation tools like navigation meshes. For object interactions, users or automated planners can place constraints, such as hand positions, on objects to encourage \model to generate pick and place motions that are plausible for a specific object size. To generate large scale motion datasets, constraints can be randomized to ensure diversity. For example, for creating a locomotion move set, a 2D root waypoint can be placed at varying angles from the starting location to synthesize locomotion in all different directions.

We evaluated \model trained for the SOMA skeleton on a diverse test suite of constraint-conditioned motion generation cases (see \cref{sec:experiments_setup} for details). On average, the model achieves 3.21 cm joint errors for full-body keyframes, 3.63 cm joint position errors for end-effector constraints, 6.88 deg joint rotation errors for end-effectors, and 3.63 cm root errors across waypoints and paths. Given such precise model outputs, light post-processing can be applied on generated motions to ensure they \textit{exactly} hit user constraints without severely degrading motion quality.
Note that, while we use the same evaluation protocol, these numbers are not comparable to those in \cref{sec:experiments}, where models are trained on a different character skeleton at 20 fps.

\subsection{Scaling Behavior}
\label{sec:key_results:scaling}
A key component of \model's success is scaling along several axes. 
To demonstrate this, we evaluate the model while varying data size, model size, and batch size (number of GPUs). 
We summarize key results in this section, while full details are provided in \cref{sec:experiments:scaling}.

\begin{figure}[t!]
    \centering
    \includegraphics[width=1.0\linewidth]{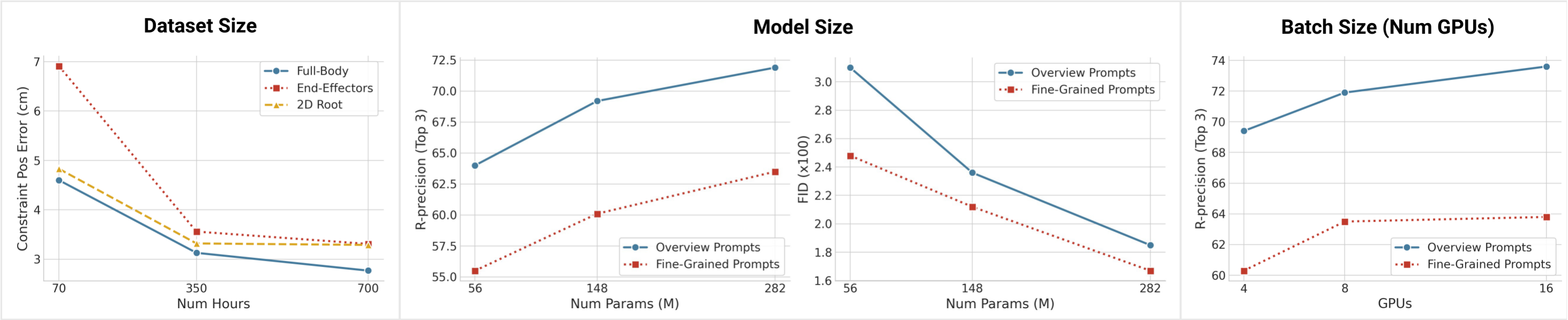}
    \caption{\textbf{Scaling Results.} Scaling dataset size, model size, and batch size improves controllability and motion quality. Increased dataset size results in greatly improved constraint following, while model size and batch size are particularly helpful for text following (R-precision) and motion quality (FID). See \cref{table:scaling} for full results.}
    \label{fig:scaling}
\end{figure}

\cref{fig:scaling} plots key metrics affected by each scaling axis.
Increasing dataset size is particularly helpful for the model to see a larger variety of motions during training, which improves its ability to follow constraints and decreases errors across all constraint types.
Increasing model size greatly improves text-following capabilities, as indicated by R-precision, along with motion quality, as indicated by FID. 
Finally, increasing the number of GPUs available for training, thereby increasing batch size, allows for further improvements particularly in text-following.

\begin{figure}[t!]
    \centering
    \includegraphics[width=1.0\linewidth]{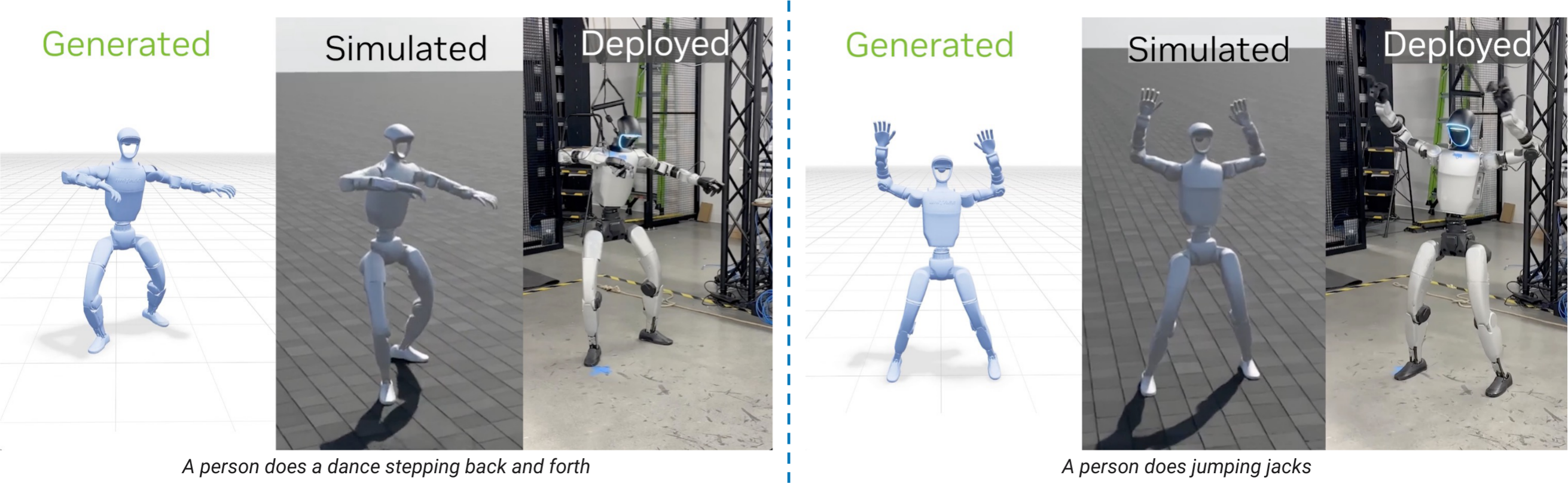}
    \caption{\textbf{Generating Robot Demonstrations.} In these results, \model is used to generate demonstration data directly on the G1 robot, which is then tracked by a physics-based humanoid policy trained with ProtoMotions~\cite{protomotions} and deployed to a real-world robot.}
    \label{fig:robot_demo}
\end{figure}

\subsection{Application: Demonstration Data for Humanoid Robots}
\label{sec:key_results:robot}

As shown in \cref{fig:robot_demo}, \model trained on G1 can directly generate demonstrations for robotics applications. Unlike traditional mocap or teleoperation, acquiring these motions takes a few seconds with a simple text prompt as input. Larger demonstration datasets can be created with batched generation and by employing constraints to ensure diversity of motion variations.

\section{Large-Scale Motion Capture Dataset}
\label{sec:dataset}
\begin{figure}[t]
    \centering
    \includegraphics[width=0.9\linewidth]{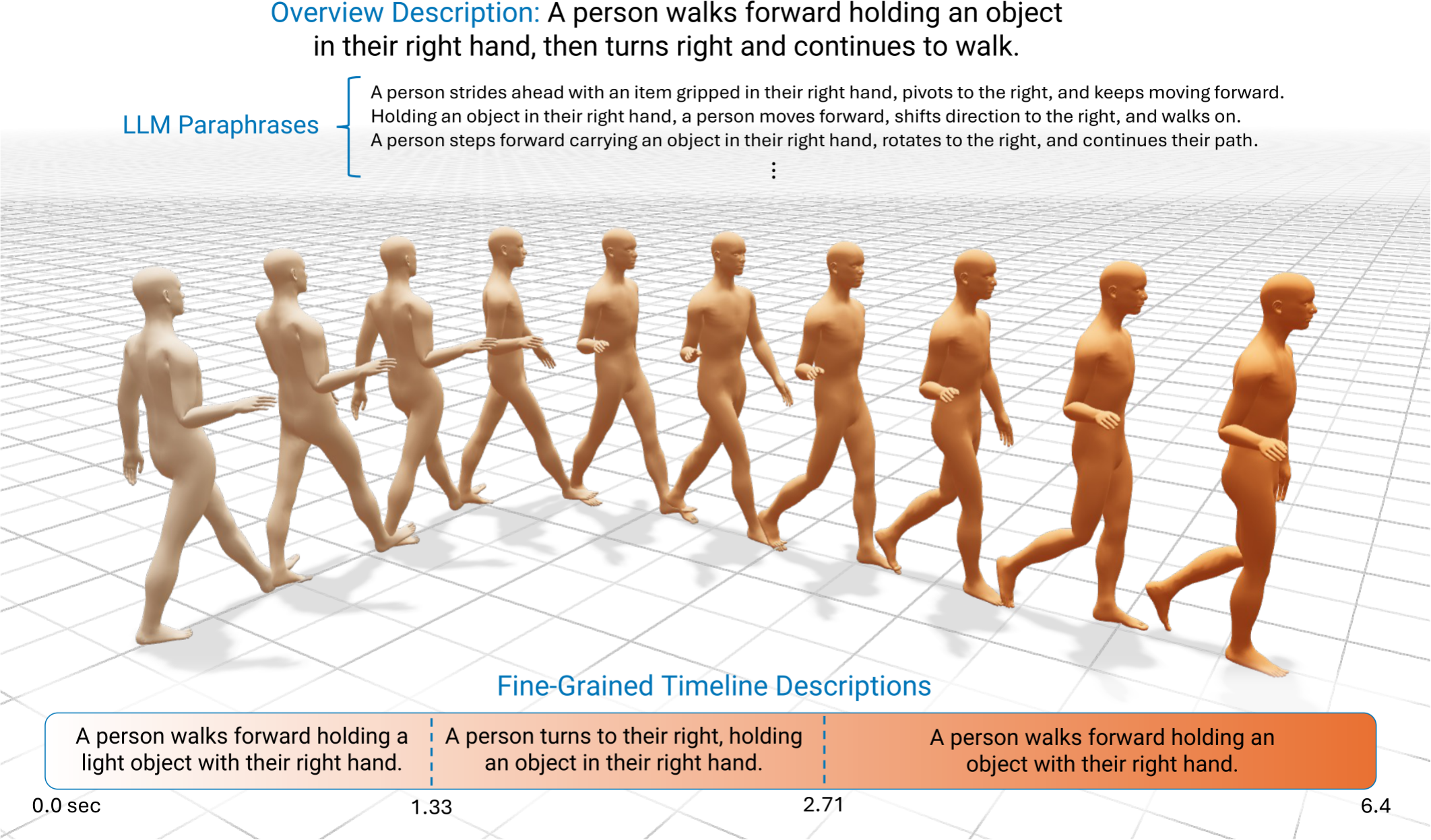}
    \caption{\textbf{Example Motion Data.} Each mocap sequence in our training dataset includes a high-level overview description, fine-grained descriptions of sub-clips containing atomic actions, and augmented LLM paraphrases.}
    \label{fig:data_sample}
\end{figure}

For training and evaluation, we leverage Bones Rigplay~\cite{bones_ai_datasets}, a large-scale optical motion capture dataset containing 700 hours of motions from 170 human subjects with a roughly equal number of male and female participants. 
The dataset contains thousands of unique actions covering a range of locomotion, gestures, everyday activities, common object interactions, videogame combat, dancing, athletics, and more. 
Many actions are also performed in different styles including tired, angry, happy, sad, scared, drunk, injured, stealthy, old, and childlike.
Actions are performed by multiple subjects across multiple takes, providing a rich diversity of performers, semantics, and motion variability. 
As shown in \cref{fig:data_sample}, each clip in the dataset is labeled with an \textit{overview} text description of the entire motion at a high level. Additionally, the clip is broken down into more \textit{fine-grained} atomic action sub-clips, each containing a text description of the contained action.

The dataset contains body-only motions (i.e., no finger motions) that have been retargeted to a uniform-proportion 27-joint skeleton to be standardized across different performers. 
We use this ``native'' 27-joint skeleton for quantitative experiments presented in \cref{sec:experiments}, however, we also retarget the dataset to other skeletons to train variations of our model, including \bodymodel~\cite{saito2026soma}, the Unitree G1 Robot~\cite{unitree2024g1}, and SMPL-X~\cite{SMPL:2015, SMPL-X:2019}. 

\paragraph{Augmentations}
For training the denoising model introduced in \cref{sec:method:diffusion_arch}, we apply augmentations to both the text and motions in the dataset. 
For text, we use an LLM (Qwen3-32B~\cite{qwen3technicalreport}) to paraphrase the motion descriptions into a consistent prompt structure (always starting with ``A [subject]...'') with various levels of detail. 
To improve handling complex prompts that contain a composition of multiple actions, we also augment the motion in the dataset by stitching together random pairs of motion clips. To ensure natural transitions between the stitched clips, we use our diffusion model trained on the non-augmented dataset to generate short transition motions between the clips. 

During training, we randomly sample from the available data and augmentation variations according to a pre-specified distribution. In particular, we train on a combination of full motion clips, single or combined action sub-clips, augmented stitched motion clips, original text descriptions, and augmented LLM paraphrases. 

\section{Method: \model}
\label{sec:method}

Our model is designed to provide users with an intuitive and versatile interface for authoring high-quality motions. 
The core component of our framework is an explicit motion diffusion model~\cite{tevet2023mdm,zhang2024motiondiffuse}, which has been shown to effectively capture the complex distribution of text and motion. 
It also enables intuitive kinematic controls through simple conditioning mechanisms, such as direct imputation of pose features~\cite{cohan2024condmdi,shafir2024human}.

\paragraph{Background}
Explicit motion diffusion~\cite{tevet2023mdm,zhang2024motiondiffuse} applies ideas from earlier image diffusion models~\cite{ho2020denoising}, but operates directly on body pose features.
Given a clean motion $\bx_0$, the forward diffusion process is defined as a Gaussian process that adds noise to the poses until they are approximately $\mathcal{N}(\mathbf{0}, \mathbf{I})$. 
To generate a motion, the reverse process is used to iteratively denoise a sequence of noisy human poses $\bx_T \sim \mathcal{N}(\mathbf{0}, \mathbf{I})$.
To enable this reverse process, a denoiser $\mathcal{D}_\theta(\bx_t, C, t)$, which takes a noisy motion $\bx_t$, conditioning signals $C$, and the current denoising step $t \in \{1, \dots, T\}$ as input, is trained to output a prediction of the clean motion $\bxhat_{0}$. The predicted clean motion is then re-noised to obtain an estimate of $\bx_{t-1}$, which serves as input to the next denoising step in the reverse diffusion process. The conditioning signals may include text or other constraints on the motion. 

The key design decisions to enable high-quality motion and control are the motion representation (\cref{sec:method:motion_rep}) and the denoiser architecture (\cref{sec:method:diffusion_arch}). In the following sections, we describe these components of our system, along with the training recipe (\cref{sec:method:training}) and generation process (\cref{sec:method:inference}).

\subsection{Motion Representation}
\label{sec:method:motion_rep}

A motion sequence is represented as $\bx = [\bx^1, \bx^2, \ldots, \bx^N]$, where N is the number of frames. We assume the motion is standardized such that the root at the first frame is above the origin. For a skeleton with $J$ joints, each pose in the motion is represented by a vector of features $[\br^p, \br^a, \bj^p, \bj^v, \bj^a, \beff]$ consisting of: 
\begin{itemize}
    \item $\br^p \in \reals^3$: \textbf{smoothed global root position}. The root motion is computed by taking the pelvis position and heavily smoothing its 2D horizontal components $(x,z)$, while keeping the $y$ height unchanged. 
    \item $\br^a \in \reals^2$: \textbf{global root heading direction}. This is represented as $[\cos(\psi), \sin(\psi)]$ with the heading angle $\psi$. The heading angle is computed based on the projection of the cross product $\mathbf{e}_y \times \mathbf{v}_\text{hips}$ onto the $xz$ (ground) plane, where $\mathbf{e}_y$ is the unit up vector and $\mathbf{v}_\text{hips}$ is the vector from left to right hip joints.
    \item $\bj^p \in \reals^{3J}$: \textbf{joint positions}. The 2D horizontal components $(x,z)$ of each joint are represented relative to the smoothed root position, while the $y$ height is global. Note that these joint positions are \textit{not} rotated to be relative to the root heading direction.
    \item $\bj^v \in \reals^{3J}$: \textbf{global joint velocities}. The joint velocities are computed from the \emph{global} joint positions, rather than the (partially) root-relative positions $\bj^p$.
    \item $\bj^a \in \reals^{6J}$: \textbf{global joint angles} encoded using the 6D representation~\cite{zhou2019continuity}.
    \item $\beff \in \{0, 1\}^4$: \textbf{foot contacts} boolean flags for the [left heel, left toe, right heel, right toe].
\end{itemize}

Considering the decomposed architecture introduced later, it is helpful to think of each pose as containing a global root component $\br^\text{glob}  = [\br^p, \br^a]$ and a body component $\bb = [\bj^p, \bj^v, \bj^a, \beff]$. 

\paragraph{Discussion}
This motion representation is carefully designed for motion quality and to be amenable for conditioning through direct imputation (\ie, overwriting) of input pose features during denoising (see \cref{sec:method:diffusion_arch}). 
\begin{wrapfigure}{r}{0.45\textwidth} %
    \centering
    \includegraphics[width=\linewidth]{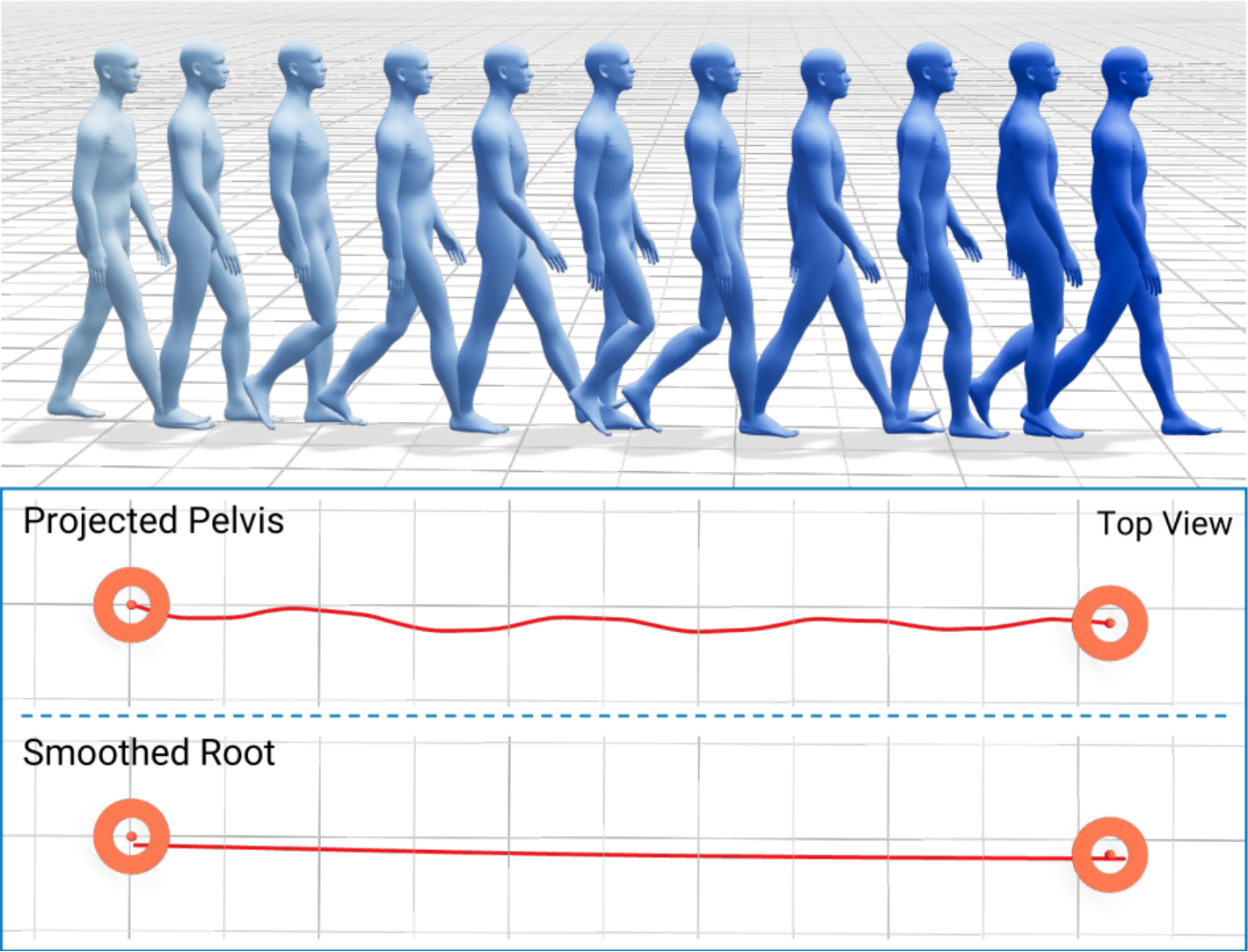}
    \caption{\textbf{Smoothed Root Representation.} For a simple walking motion, the projected pelvis path captures the sway of the hips, while the smoothed root is nearly a straight line. This smoothed trajectory offers a stable frame of reference for representing joint positions, and emulates paths drawn in practical animation tools.}
    \vspace{-5mm}
    \label{fig:smooth_root}
\end{wrapfigure}
First, our mostly global representation allows for sparse constraints on the root, joint positions, and joint rotations throughout a pose sequence. This is challenging for other common representations~\cite{guo2022humanml3d} that are purely local/velocity-based, since they require integration to determine global positions/rotations at a specific frame and are therefore better suited for temporally dense constraints~\cite{meng2025absolute}. 
Second, our joint position representation is not canonicalized with respect to the root heading in each frame. 
In prior work, joint positions are commonly canonicalized with respect to the root heading, but this design decision can lead to sudden changes in the local joint position representation due to abrupt flips in heading direction, such as during somersaults and cartwheels. These discontinuities can lead to training instabilities and compromise motion quality.
Third, using a global joint rotation representation enables users to directly specify sparse joint rotation constraints in world space (through imputation), which is rarely supported in prior systems. While this is a common requirement for animation applications, it is difficult to achieve if joint rotations are represented relative to their parent in the kinematic chain (\eg, in the SMPL body model~\cite{SMPL:2015}), since forward kinematics through the entire chain is required to recover a joint's global orientation.   
Lastly, compared to directly using the pelvis position projected to the ground as the root, our smoothed root position better emulates the smooth curves and straight lines that users are likely to specify as 2D waypoint and path constraints (\cref{fig:smooth_root}). This smoothed root representation enables our model to generate motions that follow smooth target paths, while also providing flexibility for the pelvis joint to move naturally around the smoothed path. 

\subsection{Two-Stage Transformer Denoiser}
\label{sec:method:diffusion_arch}

\begin{figure}[t]
    \centering
    \includegraphics[width=1.0\linewidth]{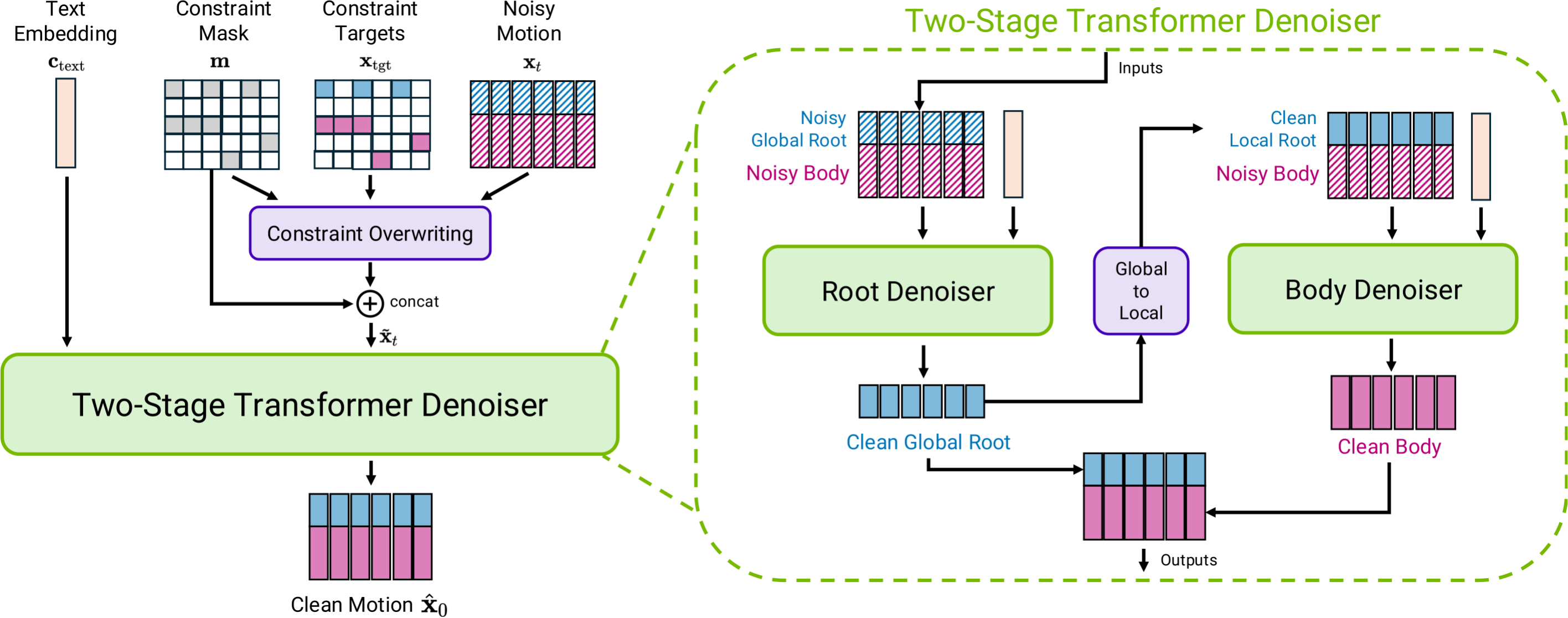}
    \caption{\textbf{Denoiser Architecture.} (Left) \model predicts clean motion given a noisy motion, pose constraints, and a text embedding. Specified pose constraints directly overwrite the noisy motion before it is given to the denoiser. (Right) The two-stage denoiser decomposes root and body motion prediction. The root denoiser first predicts the global root motion, which is transformed into a local representation as input to the body denoiser. The final output of the denoising step is the concatenation of the outputs from the two stages.}
    \label{fig:arch}
\end{figure}

Given a noisy motion $\bx_t$ at denoising step $t$, the denoiser predicts the clean motion $\bxhat_{0}$ using a two-stage transformer architecture shown in \cref{fig:arch}. Denoising is optionally conditioned on a text prompt and/or constraints.

\paragraph{Inputs}
The input noisy motion $\bx_t$ to the transformer is treated as a sequence of pose tokens. 
Kinematic constraints are specified as partial pose features in the same representation as the input motion. In particular, the target pose features are specified by a (usually sparse) target motion $\bx_{\text{tgt}}$ along with a binary control mask $\bmask$, which indicates which features are constrained. 
We impute (\ie, overwrite) the noisy motion with the target pose features according to the control mask as $\tilde{\bx}_t = \bmask \odot \bx_{\text{tgt}} + (1 - \bmask) \odot \bx_t$, where $\odot$ is the element-wise product~\cite{cohan2024condmdi}.
Finally, we concatenate the control mask with the imputed motion along the feature dimension to produce the final input tokens $\bx_{\text{in}} = [\tilde{\bx}_t; \bmask]$. 
When no constraints are specified, the mask is simply all zeros and $\tilde{\bx}_t = \bx_t$.

The denoiser prediction is $\bxhat_{0} = \mathcal{D}_\theta(\bx_{\text{in}}, C, t)$ where $C = \{ \bc_\text{text}, \bc_\text{dir}, \bc_\text{extra} \}$ is the set of additional conditioning tokens. 
$\bc_\text{text} \in \reals^{4096}$ is the LLM2Vec embedding of the input text prompt~\cite{llm2vec2024}, which we found to outperform common alternatives like CLIP~\cite{radford2021learning} and T5~\cite{raffel2020t5} in early experiments. 
$\bc_\text{dir} \in \reals^2$ is the desired heading direction of the first frame. 
Lastly, $\bc_\text{extra} \in \reals^{P \times 4096}$ contains $P$ extra all-zero tokens. 
While these extra tokens cannot exactly be considered ``register'' tokens~\cite{darcet2023vision}, since they are not learned, they achieve a similar effect of enhancing the representational capacity of the model as shown in our experiments (\cref{sec:experiments}). In practice, we use $P=49$.
All pose and conditioning tokens are embedded to the same dimensionality and added to a sinusoidal positional encoding before going into the transformer.

\paragraph{Architecture}
As shown in \cref{fig:arch}, the denoiser is decomposed into two transformer encoders: one for predicting root motion and the other for body motion.
The root denoiser first predicts the global root motion $\hat{\br}^\text{glob}_0$. Despite only predicting the root, this denoiser is conditioned on the full noisy motion $\bx_{\text{in}}$, such that the output can be coordinated with the body motion and any relevant constraints.
After the first stage, the global root prediction is transformed into a local representation $\hat{\br}^\text{local}_0$, concatenated with the body features from $\bx_{\text{in}}$, and inputted to the second transformer to predict the body motion $\hat{\bb}_0$. The final output of the denoiser is the concatenated global root and body predictions $\hat{\bx}_0 = [\hat{\br}^\text{glob}_0; \hat{\bb}_0]$.
In practice, both transformers use the same architecture with 16 layers, 8 heads, and a latent size of 1024, totaling 282 M learnable parameters in our full model.

While the global root representation is advantageous when modeling root motion, we found that using a \emph{local} root representation is more effective when conditioning the second stage of the model to denoise body motion. 
Inspired by Guo \etal~\cite{guo2022humanml3d}, we define the local root pose as $\br^\text{local} = [\dot{\br}^a, \dot{\br}^p_{xz}, \br^p_y]$ where $\dot{\br}^a \in \reals$ is the angular velocity of the heading, $\dot{\br}^p_{xz} \in \reals^2$ is the planar translation velocity of the root, and $\br^p_y \in \reals$ is the absolute height.
Converting from the global to local root representation is straightforward using finite differences to estimate velocities. 

\paragraph{Discussion}
The two-stage denoiser design is key to maximizing motion quality and control accuracy together. 
The global root representation used in the first stage enables conditioning on sparse global root constraints through direct imputation, while the second stage benefits from being conditioned on a more invariant local root representation.
Moreover, we hypothesize that body motion prediction is an easier problem when conditioned on the root motion. 
This is supported experimentally in \cref{sec:experiments}, where ablations of the two-stage model and the local root representation result in worse performance. 
Note that unlike prior two-stage approaches, our two-stage model is \emph{interleaved} since the root and body predictions are made at every denoising step. Prior work performs the root denoising process in full before generating the body motion~\cite{karunratanakul2023gmd,hwang2025motion}, while our model enables root and body corrections throughout denoising to attain better alignment in the end. 

The need for the initial heading token $\bc_\text{dir}$ as input stems from our global representation of joint rotations. 
To generalize more effectively with this global representation, we speculate that it is helpful for the model to see a wide distribution of rotations during training. Therefore, we choose not to canonicalize motions relative to the heading at the first frame, as is common practice, and instead apply a random first-frame heading augmentation to motions during training (see \cref{sec:method:training}). 
This means the model can generate motion that starts at any arbitrary heading, which we aim to control at test time with $\bc_\text{dir}$. 
The choice not to canonicalize the motion and use $\bc_\text{dir}$ is also practically convenient at test time. When the model is conditioned on global constraints in the scene or the user wants to generate motion from a sequence of multiple prompts (\eg, from an editing timeline), there is no need to apply a canonicalizing transform before generation and subsequently an inverse transform afterwards, since we can simply specify the desired initial heading with $\bc_\text{dir}$.

\subsection{Training}
\label{sec:method:training}

The denoiser is trained according to the DDPM~\cite{ho2020denoising} framework using a modified version of the simplified loss function. At each training iteration, a ground truth motion $\bx_0$ is sampled from the dataset and a diffusion timestep $t \sim \mathcal{U}\{1,\dots,T\}$ and Gaussian noise $\epsilon \sim \mathcal{N}(\mathbf{0}, \mathbf{I})$ are used to noise the motion to $\mathbf{x}_t = q(\mathbf{x}_t \mid \mathbf{x}_0, t)$. After obtaining the denoiser prediction $\hat{\bx}_0$, the loss is computed on each component of the motion representation:
\begin{align}
    \mathcal{L} &= \gamma_1 || \hat{\br}^p_0 - \br^p_0||_1 + \gamma_2 || \hat{\br}^a_0 - \br^a_0||_1 + \gamma_3 || \hat{\bj}^p_0 - \bj^p_0||_1 + \gamma_4 || \hat{\bj}^v_0 - \bj^v_0||_1 \notag \\
    &+ \gamma_5 || \hat{\bj}^a_0 - \bj^a_0||_1 + \gamma_6 || \hat{\beff}_0 - \beff_0||_1 + \gamma_7 || \text{FK}(\hat{\bj}^a_0) - \bj^p_0||_1.
\end{align}
where $||\cdot||_1$ is a smooth L1 loss that uses an L2 term when the loss is small and an L1 term otherwise~\cite{girshick2015fast}. $\text{FK}(\cdot)$ is the forward kinematics function, which computes joint positions from joint rotations. 
In practice, the losses are weighted as $\gamma_1 = \gamma_3 = \gamma_5 = 10.0$, $\gamma_2 = 2.0$, $\gamma_4 = 3.0$, $\gamma_6 = 4.0$, and $\gamma_7 = 5.0$. 
Training uses variable length sequences within each batch, so loss functions are masked accordingly. %
We use $T=1000$ diffusion steps for training.

To improve training stability, we use the Adam-atan2 optimizer~\cite{everett2024scaling} with a learning rate of $2e{-}5$. 
Ground truth motions are cropped to a maximum length of 10 sec and translated such that the root position is above the origin at the first frame. The heading direction at the first frame is randomized to ensure the model is robust to variations in global rotations. 
Our best model configuration is trained with a batch size of 2048 across 16 NVIDIA A100 (SXM4-80GB) GPUs and generates motions at 30 fps.

\paragraph{Training Curriculum}
The denoiser is trained in two phases. For the first 500k steps (phase 1), the model is trained purely on the text-to-motion task with no constraints given as input. For the second 500k steps (phase 2), the model is trained on a mix of text and kinematic constraints.
During phase 2, kinematic constraints are randomly sampled from a set of pre-defined constraint patterns designed to enable specific functionality at test time. These include: full-body joint positions at sparse keyframes, random subsets of hands and feet positions/rotations at sparse keyframes, 2D root position/heading at sparse keyframes, 2D root position/heading on dense paths, and foot contact configuration at sparse keyframes. Two constraint patterns are mixed together $25\%$ of the time, and $10\%$ of the time no constraints are used (leaving only text input).
During phase 2, the maximum number of keyframes sampled for sparse constraints increases linearly from 1 to 20, and sampling is biased towards fewer keyframes to reflect common real-world use cases. 
Dropout with a rate of 0.1 is used during phase 1, but is removed for phase 2 to avoid dropping out conditioning constraints that are directly overwritten to the noisy motion input.
During both phases, the text input is dropped $10\%$ of the time to enable classifier-free guidance at test time~\cite{ho2022classifier}. 
Exponential Moving Average (EMA) is applied every 10 steps throughout training with a decay of 0.995 to maintain an average of the denoiser parameters, which is then used at test time.

\subsection{Motion Generation}
\label{sec:method:inference}
After training, motions are generated using the DDIM~\cite{song2020denoising} inference process with, by default, 100 denoising steps.
We leverage a classifier-free guidance approach that decomposes text and constraint conditioning to allow control over each one individually. 
In particular, the model output at each denoising step is computed as $\hat{\bx}_0 = \mathcal{D}_\varnothing + w_\text{text}(\mathcal{D}_\text{text}  - \mathcal{D}_\varnothing) + w_\text{constr}(\mathcal{D}_\text{constr}  - \mathcal{D}_\varnothing)$ where $\mathcal{D}_\varnothing$ is the model output using no text or constraint conditioning, $\mathcal{D}_\text{text}$ uses only text conditioning (no constraints), and $\mathcal{D}_\text{constr}$ uses only constraint conditioning (no text). 
By default, we use $w_\text{text} = 2$ and $w_\text{constr} = 2$, but a user can adjust each to vary the influence of text and constraint conditioning on the model output. 
Several prior works make use of gradient-based guidance to further improve constraint following at test-time~\cite{rempeluo2023tracepace,karunratanakul2023gmd}, but we found that since our model is already directly conditioned on the constraints, adding gradient-based guidance gave minimal improvement, substantially increased generation time, and was generally unstable and difficult to tune.

\paragraph{Multi-Prompt Sequencing}
While \model is trained to take one text prompt as input, it is practically desirable to generate motions for a sequence of prompts (see \cref{sec:key_results:text}). 
We achieve this by generating the motion for each prompt in sequence and adding constraints to maintain plausible transitions. 
After generating motion for the first prompt, the subsequent prompt is generated with an overlap to the first prompt where several full-body keyframe constraints are added to encourage consistent joint positions and accelerations with the previously generated motion. To ensure a smooth transition, we blend the constrained frames that are shared by the first and second prompt after generation.

\paragraph{Motion Post-Processing}
In practice, post-processing can be performed on the model outputs to improve the generated motion. 
Simple foot locking and IK can clean up any undesirable foot skate using the foot contact classification directly from the model output.
It is also helpful to perform a short optimization on the output motion to ensure it \emph{exactly} hits the kinematic constraints, which is challenging for the model to achieve. We use these post-processing approaches in our released demo and codebase, but not for the experiments in \cref{sec:experiments} to ensure a fair comparison between methods.
\section{Related Work}
\label{sec:relwork}

Our approach is closely related to previous works in generative human motion modeling, while improving upon them in key ways.
Besides explicit motion diffusion~\cite{tevet2023mdm,zhang2024motiondiffuse}, several alternative approaches have proven successful at text-conditioned human motion generation. 
Latent motion diffusion~\cite{chen2023executing} denoises motion in a learned latent space rather than the explicit pose space, thereby improving efficiency. In a similar spirit, MMM~\cite{pinyoanuntapong2024mmm} and MoMask~\cite{guo2024momask} learn a discretized latent space via a VQ-VAE~\cite{van2017neural}, then train a model to generate a sequence of latents through a progressive masked prediction procedure. A different class of methods treat discretized latents as tokens and autoregressively predict them in sequence, similar to large language models~\cite{zhang2023generating,jiang2024motiongpt}. 
Such latent approaches focus primarily on text control. Those that do handle kinematic constraints require latent test-time optimization to achieve high accuracy~\cite{wan2024tlcontrol,pinyoanuntapong2025MaskControl}.

Our choice of explicit motion diffusion is motivated by the ease of controllability on the pose features.
OmniControl~\cite{xie2024omnicontrol} demonstrated dense and sparse positional control with a ControlNet~\cite{zhang2023adding} fine-tuned on top of a base motion diffusion model.
GMD~\cite{karunratanakul2023gmd} uses a combination of imputation and test-time guidance to handle potentially sparse positional constraints. Learning an RL policy on top of a diffusion model has also been explored for kinematic controls~\cite{shi2024amdm,zhao2025DartControl}.
Our method is most related to methods that use imputation to condition generation by directly overwriting constraints in the input motion~\cite{shafir2024human,hwang2025motion}.
CondMDI uses the same imputation approach that we do with a concatenated mask as input to the denoiser~\cite{cohan2024condmdi}.

Our method, \model, supports a suite of kinematic controls that is more extensive than prior works. Through direct imputation during training, our denoiser can handle both sparse and dense constraints on positions and joint rotations. This is achieved without additional ControlNet fine-tuning, test-time guidance/optimization, or RL. Moreover, our smoothed root representation lends itself to root constraints commonly found in motion editing applications.
While prior work has leveraged a global joint position representation to handle sparse constraints~\cite{karunratanakul2023gmd,cohan2024condmdi,hwang2025motion,meng2025absolute}, we also adopt the global representation for joint rotations, enabling sparse rotation control.
Additionally, previous approaches that use a two-stage model~\cite{karunratanakul2023gmd,hwang2025motion} train each stage independently, while our interleaved two-stage denoiser trains end-to-end.

A major challenge in learning effective motion generation models is the relative lack of large-scale datasets, with the most common benchmark HumanML3D~\cite{guo2022humanml3d} containing only 30 hours of motion.
Several works try to address this through increasingly larger datasets.
Motion-X was an early work to collect a dataset with a majority of motions recovered from online video sources, totaling 144 hours of motion~\cite{lin2023motion}. Datasets have continued to grow through a combination of mocap and videos from several sources~\cite{lu2025scamo,lin2025quest,wang2025scaling,zhang2024large,hymotion2025}, with the recent MotionMillion dataset containing 2000 hours of motions~\cite{fan2025gotozero}. 
While such diverse data is exciting to study the scaling properties of motion generation, these datasets rely heavily on motions reconstructed from monocular videos and/or mocap from a variety of sources with different framerates and skeletons. Additionally, text descriptions are often labeled through automatic LLM-driven approaches.
As a result, the average motion and text quality is considerably lower than the 700 hours of optical mocap data and human-labeled text annotations that \model is trained on. An exciting future direction is how to leverage large-scale video data without compromising the motion quality that can be learned from mocap.

\section{Quantitative Evaluation}
\label{sec:experiments}

In this section, we perform a detailed experimental analysis of key design decisions of our approach (\cref{sec:experiments:ablations}) and scaling behavior in terms of data size, model size, and batch size (\cref{sec:experiments:scaling}). 

Most prior works evaluate on the public HumanML3D benchmark~\cite{guo2022humanml3d}, which is relatively small scale and is becoming increasingly saturated, as indicated by better-than-ground-truth metrics reported for recent methods. 
For our experiments, we choose to exclusively use the large-scale, high-quality Bones Rigplay~\cite{bones_ai_datasets} mocap dataset described in \cref{sec:dataset}, as it provides a robust benchmark for highlighting differences between ablations and variations of our method.

\subsection{Experiment Setting}
\label{sec:experiments_setup}

\paragraph{Dataset}
We hold out 10\% of the motions in Rigplay from training for evaluation. 
Splits are determined based on unique behaviors (\ie, action types) in the dataset, such that the test set contains only novel behaviors that were not seen during training. 
For results reported here, we evaluate on a subset containing about 5k motions that cover all distinct behaviors in the test set. 
For these experiments, models are trained on the native 27-joint skeleton version of the dataset.

\paragraph{Test Cases}
Models are evaluated in three different settings. 
(1) To evaluate the text-to-motion task with no kinematic constraints, we prompt the model with the high-level \textit{Overview Prompt} for each motion (see \cref{sec:dataset}). 
(2) We also evaluate unconstrained text-to-motion with a \textit{Fine-Grained Prompt} that tends to describe a shorter motion containing an atomic action. 
(3) To evaluate combined text+constraint conditioning, we aggregate results from a test suite containing diverse variations of kinematic constraints sampled from the ground truth motion. Constraints in this suite reflect real-world use cases including sparse full-body joint position keyframes, in-betweening with blocks of keyframes at the start and/or end of a motion, sparse keyframes on position/rotation of end-effectors (hands and feet), sparse 2D waypoint keyframes for the root, dense 2D root paths, and mixing multiple constraints together. Constrained generation is evaluated at motion lengths from 3 to 9 sec, both with and without text-conditioning. For a subset of test cases, we apply random perturbations on the global translation and heading of constraints to evaluate generalization.

\paragraph{Evaluation Metrics}
We adapt common metrics from prior work~\cite{guo2022humanml3d}. 
To evaluate text-following, we report \textit{Top-3 R-precision (R@3)} using a TMR embedding model~\cite{petrovich23tmr} trained on the full Rigplay dataset, including both train and test split.  
Notably, retrieval with TMR is performed over the entire test set rather than within small batches of 32 as is common in prior work, significantly increasing the challenge.
For motion quality, we measure \textit{FID} using the same TMR model. We also report a \textit{Foot Skate} metric that computes the mean velocity of feet joints in the output motion for frames where the model predicts the joint should be in static contact with the ground. This foot skate metric depends on the \textit{Foot Contact Classification Accuracy}, which is generally very good across all methods, but is also reported for completeness. 

For constraint-conditioned generation, the \textit{average distance error} between the input constraint and the generated motion at constrained frames is reported. Errors are split into full-body positions, end-effector position/rotation, and 2D root position and averaged over all test cases.
Because our model uses a smoothed root representation, the model can generate motion where the smoothed root matches the constraint, but the projected pelvis of the character still deviates from the smooth path/waypoint constraint. As discussed in \cref{sec:method:motion_rep}, this flexibility is useful for maintaining natural motion when following root constraints, however, significant deviation of the pelvis from the constraint is undesirable as the character can appear to drift from the constraint path. 
To evaluate the extent of this deviation, we report the \textit{95th percentile of the 2D position error} between the smoothed root constraint and projected pelvis. This value indicates the maximum that the pelvis will stray from smoothed root constraints for the vast majority of generated motions; closest to ground truth is optimal.

\paragraph{Model Setting}
Unless otherwise noted, the models presented in these experiments are trained at 20 fps using 8 GPUs (resulting in a batch size of 1024 motions) and on data with all the augmentations described in \cref{sec:dataset}. 
This differs slightly from our best models demonstrated in \cref{sec:key_results}, which are 30 fps and trained with 16 GPUs, but the trends are still informative.

\begin{table*}[t]
\centering
\resizebox{\textwidth}{!}{%
\begin{tabular}{l|cccc|cccc|ccccc}
\toprule
& \multicolumn{8}{c|}{\textbf{Text-Following Evaluation}} & \multicolumn{5}{c}{\textbf{Constrained Evaluation}} \\
& \multicolumn{4}{c|}{Overview Prompt Test Set} &  \multicolumn{4}{c|}{Fine-Grained Prompt Test Set} & Full-Body & \multicolumn{2}{c}{End-Effector Joints} & 2D Root & 2D Pelvis \\
\textbf{Method} & \textbf{R@3}$\uparrow$ & \textbf{FID} $\downarrow$ & \textbf{Skate} (cm/s)$\downarrow$ & \textbf{Contact}$\uparrow$ & \textbf{R@3}$\uparrow$ & \textbf{FID}$\downarrow$ & \textbf{Skate} (cm/s)$\downarrow$ & \textbf{Contact}$\uparrow$ & \textbf{Pos} (cm)$\downarrow$ & \textbf{Pos} (cm)$\downarrow$ & \textbf{Rot} (deg)$\downarrow$ & \textbf{Pos} (cm)$\downarrow$ & \textbf{Pos@95\%} (cm) \\
\midrule
Ground Truth & 75.6 & 0.0 & 2.21 & 1.00 & 79.4 & 0.00 & 2.23 & 1.00 & - & - & - & - & 6.3 \\
\midrule
Full Model (Ours) & \textbf{71.9} & 1.85 & \textbf{3.87} & \textbf{0.98} & 63.5 & 1.67 & \textbf{3.88} & \textbf{0.98} & 2.67 & 3.09 & 4.18 & 2.90 & 9.7 \\
\midrule
One-Stage Arch & 71.5 & \textbf{1.65} & 7.59 & 0.94 & 63.5 & \textbf{1.51} & 6.80 & 0.95 & 8.37 & 10.19 & 5.19 & 7.74 & 21.1 \\
Second Stage Global & 70.3 & 1.87 & 4.17 & \textbf{0.98} & 63.2 & 1.66 & 4.07 & \textbf{0.98} & 2.97 & 3.39 & 5.67 & 3.25 & 10.2 \\
No Smoothed Root & 71.6 & 1.75 & 4.39 & 0.97 & \textbf{64.0} & 1.55 & 4.27 & \textbf{0.98} & 2.68 & 3.19 & \textbf{3.93} & 3.21 & 7.9 \\
No Extra Tokens & 70.9 & 1.95 & 4.28 & 0.97 & 61.6 & 1.77 & 4.17 & \textbf{0.98} & \textbf{2.40} & \textbf{2.59} & 5.55 & \textbf{2.85} & 9.18 \\
No Train Curriculum & 71.3 & 1.84 & 3.92 & \textbf{0.98} & 63.2 & 1.66 & 3.91 & \textbf{0.98} & 5.80 & 6.59 & 4.34 & 5.71 & 15.5 \\
\bottomrule
\end{tabular}%
}
\caption{\textbf{Ablation Study.} Evaluation of text and constraint-conditioned motion generation on the Rigplay test set. The full model is compared to various baselines to justify key design decisions, including the two-stage denoiser, smoothed root representation, and dual-phase training curriculum. All models are trained using a medium batch size (8 GPU) at 20 fps. FID is multiplied $\times100$ for readability.}
\label{table:ablation_study}
\end{table*}

\subsection{Ablation Study}
\label{sec:experiments:ablations}

In \cref{table:ablation_study}, we compare the full model architecture and training curriculum to several strong baselines to justify key design decisions. Metrics computed on the ground truth data are shown for reference.

\paragraph{Two-Stage Denoiser}
We first compare our decomposed two-stage denoiser design to a \textit{One-Stage} baseline that uses a single transformer to simultaneously denoise root and body motion. To ensure a fair comparison, we increase the number of layers and latent size of the baseline such that the number of learnable parameters is similar to the two-stage model. 
While the text-following capabilities of this baseline are about the same as the full model, we note a substantial increase in foot skating, indicating that generating body motion conditioned on root motion is indeed easier than generating both simultaneously. The one-stage baseline also causes a substantial increase in constraint errors.

The \textit{Second Stage Global} baseline uses the global root representation in the second (body) stage of the denoiser instead of converting the input root motion to a local representation. This tends to cause an increase in foot skating, likely due to the lack of invariance of the global root representation.

\paragraph{Smoothed Root Representation}
The next baseline directly uses the pelvis joint projected to the ground as the root instead of the smoothed representation used in the full model, which causes notably increased foot skate.
Without the smoothed root, body joint positions are represented with respect to the pelvis, which may be more difficult to learn due to the high-frequency motions of the pelvis.
Despite this foot skate, constraint accuracy in on par with the smoothed root representation.
As expected, the 95th percentile of pelvis error is lower than the full model, since the baseline is directly trained to match the pelvis to the root constraint. 
However, in practice this can cause qualitatively unnatural motions when constraining motions with straight lines or smoothed curves as is common in animation applications. For example, asking the baseline to generate a walking motion along a straight line results in a motion with a stealthy/sneaking style or old person style, since these minimize lateral pelvis movements and therefore deviate less from the straight line constraint.

\paragraph{Extra ``Register" Tokens}
The \textit{No Extra Tokens} baseline removes the extra register tokens, leaving only the text embedding and heading token as additional conditioning for the model. This tends to reduce performance particularly for text-following and motion quality, indicating that the added tokens increase the representational capacity of the model.

\paragraph{Dual-Phase Training Curriculum}
The \textit{No Train Curriculum} baseline directly trains the model for text+constraint conditioning from scratch, rather than training in two phases, as described in \cref{sec:method:training}. For a fair comparison, the baseline is trained for 1 million steps, the same as the total steps of phased training. During training, the baseline receives text-only or text+constraint inputs with equal probability and does not use dropout. 
This baseline reaches comparable text-following accuracy and motion quality as using phased training, but sees an increase in constraint errors. 
In two-phase training, phase 1 is dedicated to pre-training on text-to-motion so that phase 2 can be dedicated to constraint-following, but non-phased training must balance text and constraint learning for the entire duration of training. While it is possible there is a perfect balance between text-only and text+constraint input sampling that will result in competitive performance across the board, we find the phased training works well without additional hyperparameter tuning.

\begin{table*}[t]
\centering
\resizebox{\textwidth}{!}{%
\begin{tabular}{l|cccc|cccc|ccccc}
\toprule
& \multicolumn{8}{c|}{\textbf{Text-Following Evaluation}} & \multicolumn{5}{c}{\textbf{Constrained Evaluation}} \\
& \multicolumn{4}{c|}{Overview Prompt Test Set} &  \multicolumn{4}{c|}{Fine-Grained Prompt Test Set} & Full-Body & \multicolumn{2}{c}{End-Effector Joints} & 2D Root & 2D Pelvis \\
\textbf{Method} & \textbf{R@3}$\uparrow$ & \textbf{FID} $\downarrow$ & \textbf{Skate} (cm/s)$\downarrow$ & \textbf{Contact}$\uparrow$ & \textbf{R@3}$\uparrow$ & \textbf{FID}$\downarrow$ & \textbf{Skate} (cm/s)$\downarrow$ & \textbf{Contact}$\uparrow$ & \textbf{Pos} (cm)$\downarrow$ & \textbf{Pos} (cm)$\downarrow$ & \textbf{Rot} (deg)$\downarrow$ & \textbf{Pos} (cm)$\downarrow$ & \textbf{Pos@95\%} (cm) \\
\midrule
Ground Truth & 75.6 & 0.00 & 2.21 & 1.00 & 79.4 & 0.00 & 2.23 & 1.00 & - & - & - & - & 6.3 \\
\midrule
Full Dataset & \textbf{71.5} & 1.84 & \textbf{4.23} & \textbf{0.97} & 63.0 & 1.07 & \textbf{4.15} & \textbf{0.98} & \textbf{2.77} & \textbf{3.31} & \textbf{5.36} & \textbf{3.29} & 10.7 \\
50\% Dataset & 70.8 & \textbf{1.81} & 4.43 & \textbf{0.97} & \textbf{63.4} & \textbf{1.06} & 4.27 & 0.97 & 3.13 & 3.56 & 6.29 & 3.32 & 10.7 \\
10\% Dataset & 71.0 & 2.07 & 5.28 & 0.96 & 62.4 & 1.41 & 5.12 & 0.97 & 4.60 & 6.91 & 10.03 & 4.83 & 15.5 \\
\midrule
L Model (282 M) & \textbf{71.9} & \textbf{1.85} & \textbf{3.87} & \textbf{0.98} & \textbf{63.5} & \textbf{1.67} & \textbf{3.88} & \textbf{0.98} & \textbf{2.67} & \textbf{3.09} & \textbf{4.18} & \textbf{2.90} & 9.7 \\
M Model (148 M) & 69.2 & 2.36 & 4.45 & 0.97 & 60.1 & 2.12 & 4.31 & 0.97 & 3.26 & 3.72 & 4.70 & 3.34 & 10.0 \\
S Model (56 M) & 64.0 & 3.10 & 4.53 & 0.97 & 55.5 & 2.48 & 4.47 & 0.97 & 3.56 & 3.98 & 11.27 & 3.49 & 10.8 \\
\midrule
L Batches (16 GPU) & \textbf{73.6} & \textbf{1.61} & 3.97 & \textbf{0.98} & \textbf{63.8} & \textbf{1.52} & \textbf{3.87} & \textbf{0.98} & \textbf{2.33} & \textbf{2.71} & \textbf{4.09} & \textbf{2.35} & 8.5 \\
M Batches (8 GPU) & 71.9 & 1.85 & \textbf{3.87} & \textbf{0.98} & 63.5 & 1.67 & 3.88 & \textbf{0.98} & 2.67 & 3.09 & 4.18 & 2.90 & 9.7 \\
S Batches (4 GPU) & 69.4 & 2.01 & 4.45 & 0.97 & 60.3 & 1.98 & 4.27 & \textbf{0.98} & 2.97 & 3.68 & 5.61 & 3.42 & 10.1 \\
\bottomrule
\end{tabular}%
}
\caption{\textbf{Scaling Analysis.} Evaluation of text and constraint-conditioned motion generation on the Rigplay test set. (Top) Increasing the amount of training data improves motion quality and constraint accuracy due to increased diversity. (Middle) Increased model size improves performance on all metrics. (Bottom) Increasing batch size by using more GPUs generally improves performance across the board.}
\label{table:scaling}
\end{table*}

\subsection{Scaling Analysis}
\label{sec:experiments:scaling}

\cref{table:scaling} evaluates how scaling affects model performance across three different axes.

\paragraph{Data Size} 
The top part of the table compares using the full training dataset to training on a subset of 50\% and 10\% of training motions. 
Subsets are strategically sampled to include all unique behaviors (action types) from the full dataset, but with the number of performances for each behavior reduced to the desired fraction. As a result, this experiment primarily evaluates how much repeated performances of the same behavior across many actors influences motion generation ability. 
For this experiment only, we do not use the augmented stitched motions described in \cref{sec:dataset}.

Comparing the three dataset sizes, we see that foot skate and constraint accuracy monotonically improve with more available training data, with significantly decreased performance when using 10\% of the data (\ie, the same order of magnitude as popular AMASS~\cite{mahmood2019amass} and HumanML3D datasets). 
Curiously, R-precision and FID are not significantly affected by dataset size, which we believe is an artifact of how we subsample the training data. Since the baselines are trained on the same unique behaviors as the full training set, they should still generate reasonable results for test prompts. Therefore, retrieval with TMR is still successful and R-precision is not greatly affected. 
Similarly, the generated motion distributions for the test set will be similar even if using only 10\% of the data, especially after embedding with TMR and being fit with a unimodal Gaussian, as is done in the FID metric. 
What R-precision and FID do not capture are the fine-grained differences in motion distribution and subtle but important variations in motion, which manifests as reduced constraint following accuracy due to training on less diverse data.

\paragraph{Model Size}
The middle part of \cref{table:scaling} compares large (L), medium (M), and small (S) variants of our model in terms of learnable parameters. Our full best model uses the large size. The medium variant uses 8 layers in the transformer encoders instead of 16, while the small variant additionally decreases the latent size to 512 from 1024. We see that increasing model size improves performance across all metrics. While continuing to increase model size to the order of 500M or 1B parameters could potentially further improve performance, we found training stability becomes a bigger challenge and we speculate that without more data there will be diminishing returns.

\paragraph{Batch Size}
The bottom part of the table evaluates how increasing batch size affects performance. In practice, a larger batch size means using more GPUs so we compare small (S), medium (M), and large (L) batch sizes using 4, 8, and 16 GPUs, respectively. This corresponds to 512, 1024, and 2048 batch sizes.
As shown in the table, using more GPUs generally improves performance across the board as the gradient estimate during optimization becomes more accurate. 
We use the large batch size with 16 GPUs to train our best model.

\section{Conclusion}

We have introduced \model, a kinematic motion diffusion model trained on a large-scale motion capture dataset that generates high-quality human motions and can be controlled through text and a variety of kinematic constraints. 
Our model enables easy authoring of motions using intuitive interfaces as shown in \cref{sec:key_results}, and can be applied directly to generating robot motion after retargeting the training dataset. 
After training entirely on optical mocap data, our model gives a strong foundation for further scaling up of human motion generation. 

\paragraph{Future Challenges}
Looking forward, a promising direction is to further scale up the model with motions reconstructed from internet videos or generated videos. An important challenge here will be how to combine clean and noisy data sources without compromising output motion quality from the model. 
While \model is designed specifically for ``offline'' motion authoring and can take several seconds to generate a motion, applications such as robotics and digital twin simulation require a runtime model that dynamically controls humanoids and reacts to changing environments. To this end, an interesting avenue is moving diffusion to a learned latent space and reformulating motion generation to be an autoregressive problem. 
Finally, scene and object interactions are crucial to making motion generation models truly practical for most applications. Gathering data for this problem becomes even more challenging, and will require creative solutions.

\clearpage %

\setcitestyle{numbers}
\bibliographystyle{plainnat}
\bibliography{references}

\appendix
\newpage
\clearpage

\section{Acknowledgments}
We would like to thank John Malaska, Will Telford, Jon Shepard, and Anna Minx for helpful guidance throughout development. Thanks to Or Litany, Zhengyi Luo, Yeongho Seol, Jun Saito, and Michael Buttner for their insightful discussions on human motion. Thanks to Cyrus Hogg, Lindsey Pavao, Jenna Diamond, Rizwan Khan, Samantha Shinagawa, and Akanksha Shukla for their efforts on data acquisition and labeling. Thanks to Kaifeng Zhao and Sunmin Lee for research discussions and testing and feedback of the model and code.

\section{Contributors}

\begin{itemize}
    \item \textbf{Project Lead}: Davis Rempe
    \item \textbf{Model}: Mathis Petrovich, Davis Rempe, Ye Yuan, Haotian Zhang, Xue Bin (Jason) Peng, Yifeng Jiang, Tingwu Wang, Umar Iqbal, David Minor, Michael de Ruyter, Jiefeng Li, Chen Tessler
    \item \textbf{Data}: Edy Lim, Eugene Jeong, Sam Wu, Ehsan Hassani, Michael Huang, Jin-Bey Yu, Chaeyeon Chung, Lina Song, Olivier Dionne
    \item \textbf{Advising}: Sanja Fidler, Simon Yuen, Jan Kautz 
\end{itemize}

\end{document}